%% file: main.tex
\documentclass[10pt,twocolumn,letterpaper]{article}
\usepackage{filecontents}

\usepackage[pagenumbers]{cvpr} 

\usepackage[utf8]{inputenc} 
\usepackage[T1]{fontenc}   
\usepackage{url}            
\usepackage{booktabs}      
\usepackage{amsfonts}      
\usepackage{nicefrac}       
\usepackage{microtype}    
\usepackage{subcaption}
\usepackage{algorithm}
\usepackage{amsmath}
\usepackage{mathabx}
\usepackage{algorithmic}
\usepackage{wrapfig}
\usepackage{diagbox}
\usepackage[mathscr]{eucal}
\newsavebox{\bigimage}

\input{preamble}

\usepackage{xr}
\makeatletter

\newcommand*{\addFileDependency}[1]{
\typeout{(#1)}
\@addtofilelist{#1}
\IfFileExists{#1}{}{\typeout{No file #1.}}
}\makeatother

\newcommand*{\myexternaldocument}[1]{%
\externaldocument[#1-]{#1}%
\addFileDependency{#1.tex}%
\addFileDependency{#1.aux}%
}

\myexternaldocument{supplement}
\definecolor{cvprblue}{rgb}{0.21,0.49,0.74}
\usepackage[pagebackref,breaklinks,colorlinks,citecolor=cvprblue]{hyperref}

\def\paperID{11602} 
\def\confName{CVPR}
\def\confYear{2024}

\title{Training-Free Pretrained Model Merging}

\author{Zhengqi Xu$^1$, Ke Yuan$^1$, Huiqiong Wang$^2$, Yong Wang$^3$, Mingli Song$^1$, Jie Song$^{1*}$\\
$^1$Zhejiang University\\
$^2$Ningbo Innovation Center, Zhejiang University\\
$^3$State Grid Shandong Electric Power Company\\
{\tt\small \{xuzhengqi,keyuan,huiqiong\_wang,brooksong,sjie\}@zju.edu.cn,wangyong@sd.sgcc.com.cn}
}

\definecolor{imp}{rgb}{1,0,0} 
\definecolor{dec}{rgb}{0,1,0} 

\begin{document}

\maketitle
\renewcommand{\thefootnote}{\fnsymbol{footnote}}
\footnotetext[1]{Corresponding author.}%
\renewcommand{\thefootnote}{\arabic{footnote}}


\begin{abstract}
Recently, model merging
techniques have surfaced as a solution to combine multiple single-talent models
into a single multi-talent model. However, previous endeavors in this field have either necessitated additional training or fine-tuning processes, or require that the models possess the same pre-trained initialization. In this work, we identify a common drawback in prior works \textit{w.r.t.} the inconsistency of unit similarity in the weight space and the activation space.  To address this inconsistency, we propose an innovative model merging framework, coined as merging under dual-space constraints (MuDSC). Specifically, instead of solely maximizing the objective of a single space, we advocate for the exploration of permutation matrices situated in a region with a unified high similarity in the dual space, achieved through the linear combination of activation and weight similarity matrices.  
In order to enhance usability, we have also incorporated adaptations for group structure, including Multi-Head Attention and Group Normalization. Comprehensive
experimental comparisons demonstrate that MuDSC can significantly boost the performance of merged models with various task combinations and architectures. Furthermore, the visualization of the merged model within the multi-task loss landscape reveals that MuDSC enables the merged model to reside in the overlapping segment, featuring a unified lower loss for each task. Our code is publicly available at \href{https://github.com/zju-vipa/training_free_model_merging}{https://github.com/zju-vipa/training\_free\_model\_merging}.

\end{abstract}

\section{Introduction}
\label{sec:intro}
Great success has been achieved on various challenging tasks in computer vision by using deep neural networks, and a plethora of deep neural networks are developed and released publicly, with either their architectures and trained parameters~(\eg, Pytorch Hub\footnote{\url{https://pytorch.org/hub/}}, Hugging Hub\footnote{\url{https://huggingface.co/HUB}}). These off-the-shelf models are finely-tuned for a wide range of tasks, providing users with substantial convenience. However, these models are restricted to the tasks they were trained on, and this limitation poses significant challenges in terms of model storage \cite{fifty2021efficiently, zhang2021survey} and computation as the parameters of models continue to grow rapidly.

\begin{figure}[t]
  \centering
  \includegraphics[width=0.5\textwidth]{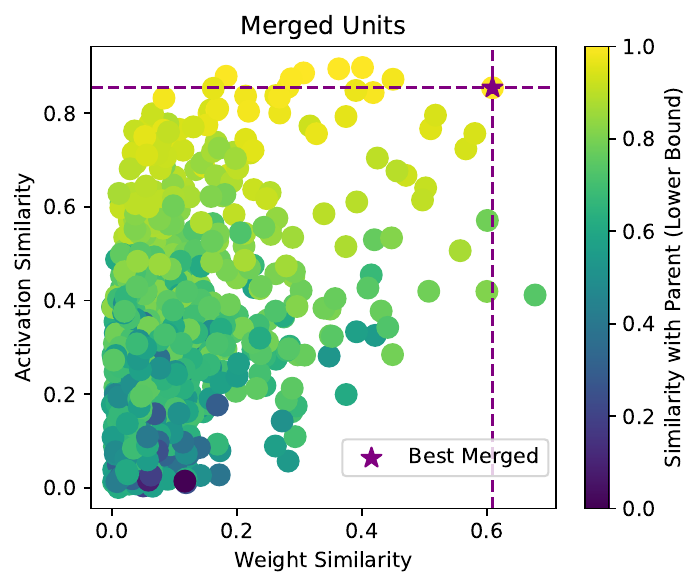}
   \caption{We use an intermediate layer of ResNet50 \cite{resnet_he2016deep} to construct two groups of units as two parents, and then merge them pairwise. In the figure, each point represents a merged unit. The x-axis and the y-axis denote the weight and activation similarity between their two parents respectively. The color indicates the smaller activation similarity between them and their two parents. The star depicts the best merged unit which is most activation-similar to its parents. More details are provided in the supplementary material.}
   \label{fig:setup_exp}
\end{figure}

Given the abundance of well-trained models across various tasks, numerous research studies propose the amalgamation of multiple models into a single model that possesses multiple functionalities simultaneously. The existing body of literature can be broadly classified into two schools: \textit{training-based knowledge amalgamation}~\cite{shen2019amalgamating,ye2019student,luo2019knowledge,yang2022factorizing,Luo2020CollaborationBC,thadajarassiri2021semi,jing2021amalgamate,carvalho2022class,gao2023contrastive,thadajarassiri2023ka} and \textit{training-free model merging}~\cite{ilharco2023editing,yadav2023ties-merging,AdaMerging_Arxiv_2023}. The former is inspired by knowledge distillation and adopts the outputs from multiple pretrained teachers to train a student model, which amalgamates the knowledge from multiple models into a single one. However, these methods usually require an additional expensive training process to facilitate the transfer of knowledge from the teacher models to the student model. In contrast, The latter fully utilize the parameters of the pretrained models to construct the multi-talent model, thereby avoiding additional training. The mainstream involves constructing multi-task models by adding task vectors, which are computed by subtracting the pre-trained model’s parameter values from those of the fine-tuned model, together. However, these methods can only merge fine-tuned models derived from the same pre-trained initialization, that is an extremely limited prerequisites. Emerging work \textit{w.r.t.} unit matching \cite{ainsworth2023git,jordan2023repair,otfusion_singh2020model,tatro2020optimizing} shows promise to breaking the extra-training and pre-training curse, which has already reached zero-loss barriers in single-task model merging \cite{ainsworth2023git}. However, the permutation invariance \cite{entezari2021role}, a cornerstone for most unit matching methods, seems not applicable to merge models from multiple tasks and leads to a poor artifact.


In this work, we have cast doubt upon the matched units identified by existing unit matching algorithms. Reviewing prior work, current unit matching methods either represent units using weights or activation values \cite{li2023deep}.
Nonetheless, using either weights or activation values as the matching criterion is incomplete. In terms of weights, the weight of a unit reflects its connection strength with the units in the previous layer, but it cannot capture the information of specific features (\textit{i.e.} scale, offset) due to the absence of input data. As shown in Fig. \ref{fig:setup_exp}, there are many pairs of units with highly similar weights but entirely different activation values, leading to a significant change in the function of their merged units. In terms of activations, it is compelling to measure units by their activations since two models must learn similar features to accomplish the same task \cite{li2015convergent}, yet the same task can be achieved by distinct parameters \cite{frankle2018lottery}.
Moreover, a phenomenon is shown in Fig. \ref{fig:setup_exp} that the best-merged unit is one whose parents simultaneously exhibits high similarity in both weight and activation.

Motivated by this, we propose a concise and effective model merging framework, termed \textit{merging under dual-space constraints(MuDSC)} to achieve a more precise matching of relevant units. Specifically, the proposed MuDSC constrains the matching algorithm to find more suitable matches by linearly combining the similarity matrix of representations of units in both activation space and weight space. For the sake of rigor, we provide a formula derivation to prove the equivalence of this approach with a linear combination of optimization objectives in separate spaces. Moreover, we devise a novel iterative algorithm that simultaneously maximize the global similarity of matched units in both weight space and activation space. In addition, we devise two variant of matching algorithms proposed in \cite{ainsworth2023git} and \cite{stoica2023zipit} to support the use of unit matching in networks with a group structure \textit{i.e.} ViT \cite{vit_dosovitskiy2021an} and Group Normalization \cite{group_norm_Wu_2018_ECCV}.


In summary, the key contributions of this paper are:
\begin{enumerate}

\item We highlight the inconsistency of unit similarity in weight space and activation space, demonstrating through a straightforward experiment the impact of this inconsistency on model merging.

\item To take into account this inconsistency, we propose a novel model merging framework MuDSC (Merging under Dual-Space Constraints) that encourages both the activation spaces and weight spaces between two models be of high similarity.


\item We present a vigorous experimental study and show that MuDSC can significantly boost the merged performance in manifold multi-task scenarios for various architectures. Further loss landscape visualization verifies that our solution locates in the overlapping part with unified lower loss of each task.

\end{enumerate}

\section{Related Work}

\noindent
\textbf{Training-based model merging.}  In order to reuse trained models
and reduce the cost of training the new models from scratch,  Shen \textit{et al.} \cite{shen2019amalgamating} propose knowledge amalgamation (KA), the goal of which is to learn a versatile student model from multiple task-specific teachers, but it is only verified on the comprehensive classification tasks. Following the knowledge amalgamation, Ye \textit{et al.} \cite{ye2019student} validate the effectiveness of KA in complex scenarios such as deep estimation and scene analysis, and in another work \cite{Ye2019AmalgamatingFK}, they customized student models by amalgamating filtered knowledge from different teachers, even surpassing the those teachers in certain cases. There are more learning strategies have been proposed to enhance the performance \cite{luo2019knowledge,shen2019customizing,Luo2020CollaborationBC} and usability \cite{ye2020data,jing2021amalgamate,zhang2023ka_obj,gao2023contrastive,jing2023deep} of KA. Although the success achieved by KA, a completely re-parameterized student model evidently leads to inefficient model merging. 
In contrast, model stitching \cite{lenc2015understanding} more efficiently leverages pre-trained models. model stitching aims to “plug-in” the bottom layers
of one network into the top layers of another network, thus forming a stitched network \cite{bansal2021revisiting,csiszarik2021similarity}. For instance, Yang \textit{et al.} \cite{yang2022deep} introduce a novel two-stage strategy for reassembling customized networks from a zoo of pre-trained models under user-specified
constraints, and Pan \textit{et al.} \cite{Pan_2023_CVPR} propose Stitchable Neural Networks for elastic deep learning by directly utilising the pretrained model families in model zoo via model stitching. Regrettably, the fine-tuning process is still necessary for bridging the gaps between pre-trained blocks in model stitching. Unlike KA and model stitching, we do not involve any additional training or fine-tuning processes in our merging methods.

\noindent
\textbf{Training-free model merging.} Considering the data privacy and the costly retraining for expanding model capabilities, recent efforts have sparked significant interest in exploring how to merge models from multiple tasks into the single multi-task model without additional training \cite{jin2023dataless, ilharco2023editing}. Ilharco \textit{et al.} \cite{ilharco2023editing} present a paradigm for editing neural networks, namely Task Arithmetic, that constructs multi-task models with performing simple addition operations on task vectors. Expanding on this groundwork, a series of work expand the scope of the task arithmetic framework \cite{zhang2023composing} or solve the potential issues of task vector-based method \cite{yadav2023ties-merging,AdaMerging_Arxiv_2023}. However, task vector-based methods can only be applied to models fine-tuned from the same pre-trained initialization, and this assumption severely restricts the applicability of model merging. Fortunately, unit matching technologies show promise in relaxing this assumption \cite{ainsworth2023git}. Guided by the existence of the permutation symmetries of hidden units in neural networks \cite{HECHTNIELSEN1990129,FUKUMIZU2000317,entezari2021role,pmlr-v202-grigsby23a,simsek2021geometry,brea2019weight,tatro2020optimizing}, unit matching aims at mapping all models to the same basin by permutating neuron units, Thought we find it widely applied for merging models in the single-task scenarios \cite{pmlr-v97-yurochkin19a,Wang2020Federated,otfusion_singh2020model,ainsworth2023git,jordan2023repair}, single-task zero-loss barrier merging \cite{ainsworth2023git} highlights the immense potential of unit matching in combining models from multiple task-specific tasks. Current unit matching methods primarily consider two types of matching spaces \cite{li2023deep}: one is the activation space\cite{stoica2023zipit,tatro2020optimizing,otfusion_singh2020model,Song_2023_ICCV,cross_layer_otfusion,benzing2022random,ainsworth2023git,li2015convergent}, and the other is the weight space \cite{Wang2020Federated,pmlr-v97-yurochkin19a,ainsworth2023git,pmlr-v189-xiao23b,pena2023re,pmlr-v139-lam21a}. Nevertheless, we demonstrate through a straightforward experiment that the similarity of units in the activation space and weight space is not always consistent, and these discrepancies manifest various pros and cons during model merging. Furthermore, simultaneously considering the similarity of models in both weight space and activation space can offer potential advantages in reducing merging barriers. Therefore, our method simultaneously constrains unit matching in both the weight space and activation space.

\section{Methodology}

In Sec. \ref{sec:preliminaries}, we introduce a general procedure of merging model via unit matching. Afterward, we provide the theoretical basis for merging models under dual-space constraints and then elaborate on how to merge models under dual-space constraints in Sec \ref{sec:mudsc}. Two solutions for adaption to group structures are presented in Sec. \ref{sec:group}.
\subsection{Preliminaries}
\label{sec:preliminaries}


Assume there are $N$ pre-trained models in the same architecture. The $n$-th model is parameterized by weights $\{\boldsymbol{W}_{l}^{(n)}\}_{l=1}^L$, where $L$ represents the number of layers of the models. Without loss of generality, we omit the bias term in parameters and simply regard the model as MLP model, so $\boldsymbol{W}_{l}^{(n)} \in \mathbb{R}^{D_lD_{l-1}}$, where $D_l$ and $D_{l-1}$ represent the number of units in $l$-th layer and $(l-1)$-th layer, respectively. 



The main idea behind model merging revolves around the fusion of units that exhibit significant similarity, aiming to maximize the cumulative global similarity of the merged units within each layer.
The unit parameters after merging are obtained by averaging the unit parameters before merging. The solution can be summarized as computing the maximum-weighted matching of the graph, where the nodes refer to the units and the edges refer to their similarities. The goal is to find a subset of edges with the maximum weight, and in which no node occurs more than $N$ times. Formally, in the $l$-th layer, let $\mathbb{C}_l \in \mathbb{R}^{(N \cdot D_l)(N \cdot D_l)}$ be the similarity matrix of merging a pair of units in the N models, where each element $(\mathbb{C}_l)_{i,j}$ denotes the similarity of the $i$-th unit and the $j$-th unit, \ie, the weights of edges as mentioned above. Then the goal of model merging can be formulated as follows:

\begin{equation}
\label{eq:objective}
\begin{aligned}
    \arg \max_{\mathbb{P}_l} &\sum_{l=1}^{L} \langle \mathbb{C}_l, \mathbb{P}_l\rangle, 
    ~\mathrm{s.t.} \sum_{i=1}^{N \cdot D_l} (\mathbb{P}_l)_{i,j} \leq N ~\forall ~j,\\ 
    &(\mathbb{P}_l)_{i, j}=1 \implies (\mathbb{P}_l)_{i, k} = (\mathbb{P}_l)_{j, k}, \forall~i, j, k\\ 
    &(\mathbb{P}_l)_{i,i}=1 ~\forall ~i.
\end{aligned}
\end{equation}
$\mathbb{P}_l$ denotes the merging matrix which is symmetrical and binary, and where $(\mathbb{P}_l)_{i,j}=1$ represents that the $i$-th unit and $j$-th unit are merged together. $\langle \boldsymbol{X},\boldsymbol{Y}\rangle=\sum_{i,j}\boldsymbol{X}_{i,j}\boldsymbol{Y}_{i,j}$ denotes the Frobenius inner product between real-valued matrices $\boldsymbol{X}$ and $\boldsymbol{Y}$. 



To merge the similar units from different models, we can construct a set of permutation matrices $\{\boldsymbol{P}_l\}_{l=1}^L$. 
Then the merge operation can be formulated as follows:
\begin{equation}
\label{eq:permutate_weight}
    {\boldsymbol{W}_l}' = \frac{1}{N}\sum_{n=1}^{N}(\boldsymbol{P}_l^{(n)})^\top {\boldsymbol{W}_l^{(n)}} \boldsymbol{P}_{l-1}^{(n)}, ~l\in\{1,2,...,L\}
\end{equation}
To clarify the relation between $\boldsymbol{P}_l^{(n)}$ and $\mathbb{P}_l$, we define $\boldsymbol{P}_l=\boldsymbol{P}_l^{(1)} \circ \boldsymbol{P}_l^{(2)} \circ ... \circ \boldsymbol{P}_l^{(N)}$ where $\boldsymbol{P}_l^{(n)} \in \mathbb{R}^{D_lD_l}$ and operation $\circ$ denotes concatenating the matrices in the first dimension. Then $\boldsymbol{P}_l$ can be straightforwardly derived from $\mathbb{P}_l$ by removing duplicate columns of $\mathbb{P}_l$ because of the constraints of Eq. (\ref{eq:objective}). 

\subsection{Merging under Dual-Space Constraints}
\label{sec:mudsc}
As introduced in Section~\ref{sec:intro}, prior works usually measure the unit similarity within a single space~(\textit{i.e.} the activation space or the weight space), resulting in inconsistencies between the two spaces. In this work, we assess unit similarity by considering similarities in both activation space and weight space. In this study, we propose an innovative merging framework that addresses disparities between models in both the activation space and weight space. The fundamental challenge lies in reconciling the distinctions between weight-based matching and activation-based matching during the matching procedure.


Given the $\{\boldsymbol{A}_l\}_{l=1}^L$ and $\{\boldsymbol{Z}_l\}_{l=1}^L$ as the representation vectors of the $l$-th layer in the activation space and the weight space, respectively, and with a balanced factor $\alpha \in [0,1]$, the expected objective can be written as follows:

\begin{equation}
\label{eq:raw_combine}
        \arg \max_{\mathbb{P}_l}  \alpha \sum_{l=1}^{L} \langle \mathbb{C}(\boldsymbol{Z}_l), \mathbb{P}_l \rangle \\ + (1-\alpha) \sum_{l=1}^{L} \langle \mathbb{C}(\boldsymbol{A}_l), \mathbb{P}_l\rangle
\end{equation}
Here, $\mathbb{C}:\mathbb{R}^{(N \cdot D_l)R_l} \rightarrow \mathbb{R}^{(N \cdot D_l)(N \cdot D_l)}$ denote the function that calculate the similarity matrix by $\boldsymbol{A}_l$ or $\boldsymbol{Z}_l$. Specially, $R_{l}$ refers to the quantity of related paramaters when it comes to $\boldsymbol{Z}_l$, or $R_{l}$ training data points when it comes to $\boldsymbol{A}_l$. We omit the constraint conditions here. Then we can re-express Eq. (\ref{eq:raw_combine}) as follows:
\begin{equation}
\label{eq:mudsc_obj}
        \arg \max_{\mathbb{P}_l}  \sum_{l=1}^{L}  \langle \alpha \mathbb{C}(\boldsymbol{Z}_l)  + (1-\alpha) \mathbb{C}(\boldsymbol{A}_l), \mathbb{P}_l \rangle
\end{equation}

Manifestly, the aforementioned derivation demonstrates that the dual-space constraints can be realized through a linear combination of the similarity matrices in both spaces. This formulation presents an efficacious approach to address unit matching under the dual-space constraints. The remaining question is how to implement Eq. (\ref{eq:mudsc_obj}) in the matching procedure. To this end, we delve into the solutions within two distinct spaces. 

In term of the activations, activation-based matching can be solved in a single-pass. This procedure is described as:


\begin{equation}
\label{eq:solving_act}
\boldsymbol{P}_l=\Psi(\mathbb{C}(\boldsymbol{A}_l))
    , l \in \{1,2,\dots,L\}
\end{equation}
Here, $\Psi(\cdot)$ is the function that derives the permutation matrix by Eq. (\ref{eq:objective}) from the similarity matrix.

In the realm of weight space, there are three distinct perspectives on how to obtain representation vectors. One viewpoint advocates the use of output vectors in the next layer to represent the units \cite{pmlr-v97-yurochkin19a}. Another approach leans towards utilizing the weights of the current layer \cite{Wang2020Federated,singh2020model}. An advanced solution involves considering all weights associated with the layer, including those from the preceding and subsequent layers. Such a method has achieved zero-loss barriers in the merging of single-task models \cite{ainsworth2023git}. In our work, we adopted the third method to obtain representations of units in the weight space. Consequently, for $n$-th model, the representation vectors in the $l$-th layer is conducted by:

\begin{equation}
\label{eq:weight_vec}
    \boldsymbol{Z}_l^{(n)} = \boldsymbol{W}_l^{(n)} \boldsymbol{P}_{l-1}^{(n)} \Vert \boldsymbol{P}_{l+1}^{(n)} (\boldsymbol{W}_{l+1}^{(n)})^\top
\end{equation}
where $\Vert$ denote the operation which concatenates the given matrices in the second dimension. Note that aforementioned $\boldsymbol{W}_l$ and $\boldsymbol{Z}_l$ contain the units of all models in the $l$-th layer like $\boldsymbol{P}_l$ and can be expressed as $\boldsymbol{W}_l^{(1)} \circ...\circ \boldsymbol{W}_l^{(N)}$ and $\boldsymbol{Z}_l^{(1)} \circ...\circ \boldsymbol{Z}_l^{(N)}$, respectively.



There is a contradiction that a valid $\boldsymbol{Z}_l$ for matching in $l$-th layer requires the units in the $(l+1)$-th and $(l-1)$-th layers to reach the optimal match. This contradiction can be resolved through an iterative algorithm, and its recursive expression is as follows:


\begin{equation}
\label{eq:solving_weight}
\begin{cases}
\boldsymbol{P}_l^{1}= \mathbb{I}
\\{\boldsymbol{Z}_l}^{t+1}= \upsilon(\boldsymbol{P}_{l-1}^{t},\boldsymbol{P}_{l+1}^{t},\boldsymbol{W}_{l}^{t},\boldsymbol{W}_{l+1}^{t})
\\\boldsymbol{P}_l^{t+1}= \Psi(\mathbb{C}({\boldsymbol{Z}_l}^{t+1}))
\end{cases},
\end{equation}
where $l \in \{1,2,\dots,L\}$ and $t \in \mathbb{N}^{+}$. 
When the number of iterations approaches infinity, the model reaches its optimal match in weight space. In Eq. (\ref{eq:solving_weight}), $\mathbb{I}$ is a permutation matrix where the submatrix of each model is identity matrix. And $\upsilon(\cdot,\cdot,\cdot,\cdot)$ is the function that applies permutation matrices to the weights of all models by Eq. (\ref{eq:weight_vec}) and then concatenate all the $\boldsymbol{Z}_l^{(n)}$ into $\boldsymbol{Z}_l$.

Afterwards, it is evident to implement the merging method under the dual-space constraints based on Eq. (\ref{eq:mudsc_obj}). We substitute Eq. (\ref{eq:mudsc_obj}) and Eq. (\ref{eq:solving_act}) into Eq. (\ref{eq:solving_weight}), yielding:


\begin{equation}
\label{eq:solving_mudsc}
\begin{cases}
\boldsymbol{P}_l^{1}= \Psi(\mathbb{C}(\boldsymbol{A}_l))
\\{\boldsymbol{Z}_l}^{t+1}= \upsilon(\boldsymbol{P}_{l-1}^{t},\boldsymbol{P}_{l+1}^{t},\boldsymbol{W}_{l}^{t},\boldsymbol{W}_{l+1}^{t})
\\\boldsymbol{P}_l^{t+1}= \Psi(\alpha\mathbb{C}({\boldsymbol{Z}_l}^{t+1})+(1-\alpha)\mathbb{C}(\boldsymbol{A}_l))
\end{cases},
\end{equation}

Lastly, we present the overall procedure of our MuDSC merging in Alg. \ref{alg:MuDSC}. We further illustrate the convergence and the complexity of Alg. \ref{alg:MuDSC} in the supplementary material.

\begin{algorithm}[t]
  
  \caption{MuDSC Merging}
  \label{alg:MuDSC}
\begin{algorithmic}
\REQUIRE  The weights $\{\boldsymbol{W}_l\}_{l=1}^{L}$ and the activations $\{\boldsymbol{A}_l\}_{l=1}^{L}$ of the models, the function for computing the similarity matrix $\mathbb{C}$, selected matching algorithm $\Psi$, the balanced factor $\alpha$. $RP$ returns a random permutation of the sequence.
\FOR{$l=1,2,\dots,L$}
    \STATE ${\boldsymbol{C}_l}' \gets \mathbb{C}(\boldsymbol{A}_l)$
    \STATE $\boldsymbol{P}_l \gets \Psi({\boldsymbol{C}_l}')$
\ENDFOR
\REPEAT
    \FOR{$l=RP(1,2,\dots,L)$}
        \STATE ${\boldsymbol{Z}_l} \gets \upsilon(\boldsymbol{P}_{l-1},\boldsymbol{P}_{l+1},\boldsymbol{W}_{l},\boldsymbol{W}_{l+1})$
        \STATE ${\boldsymbol{C}_l}' \gets \alpha \mathbb{C}(\boldsymbol{Z}_l) + (1-\alpha) {\boldsymbol{C}_l}'$
        \STATE $\boldsymbol{P}_l \gets \Psi({\boldsymbol{C}_l}')$
    \ENDFOR
\UNTIL{convergence}
\STATE Get the merged weights $\{\boldsymbol{W}_l'\}_{l=1}^{L}$ by Eq. (\ref{eq:permutate_weight})
\STATE \textbf{return} $\{\boldsymbol{W}_l'\}_{l=1}^{L}$
\end{algorithmic}
\end{algorithm}

\subsection{Adaptation to Group Structure}
\label{sec:group}
As mentioned previously, the abstractions of current matching algorithm do not generalize to networks with the following structure:

\begin{enumerate}
    \item \textbf{External Permutation Invariance}: There are $g$ separate groups which can be permuted with each other. For instance, Group Normalization \cite{group_norm_Wu_2018_ECCV} divides channels into $g$ groups and normalizes the features within each group.
    \item \textbf{Internal Permutation Invariance}: Each of those $g$ groups contains $k$ units which can be permuted. For instance, the features within each group of Group Normalization can be permuted.
\end{enumerate}

This structure is commonly encountered within contemporary neural network modules, in particularly Group Normalization \cite{group_norm_Wu_2018_ECCV} and Multi-Head Attention \cite{vaswani2017attention}. Consequently the application of matching algorithms will face significant hindrance if they cannot adapt to this structure. Here, we propose viable adaptation strategies for a alignment algorithm based on linear sum assignment \cite{ainsworth2023git} (namely group alignment) and the greedy algorithm used in zip operation \cite{stoica2023zipit} (namely group zip). 

\begin{figure}[h]
  \centering
  \includegraphics[width=0.49\textwidth]{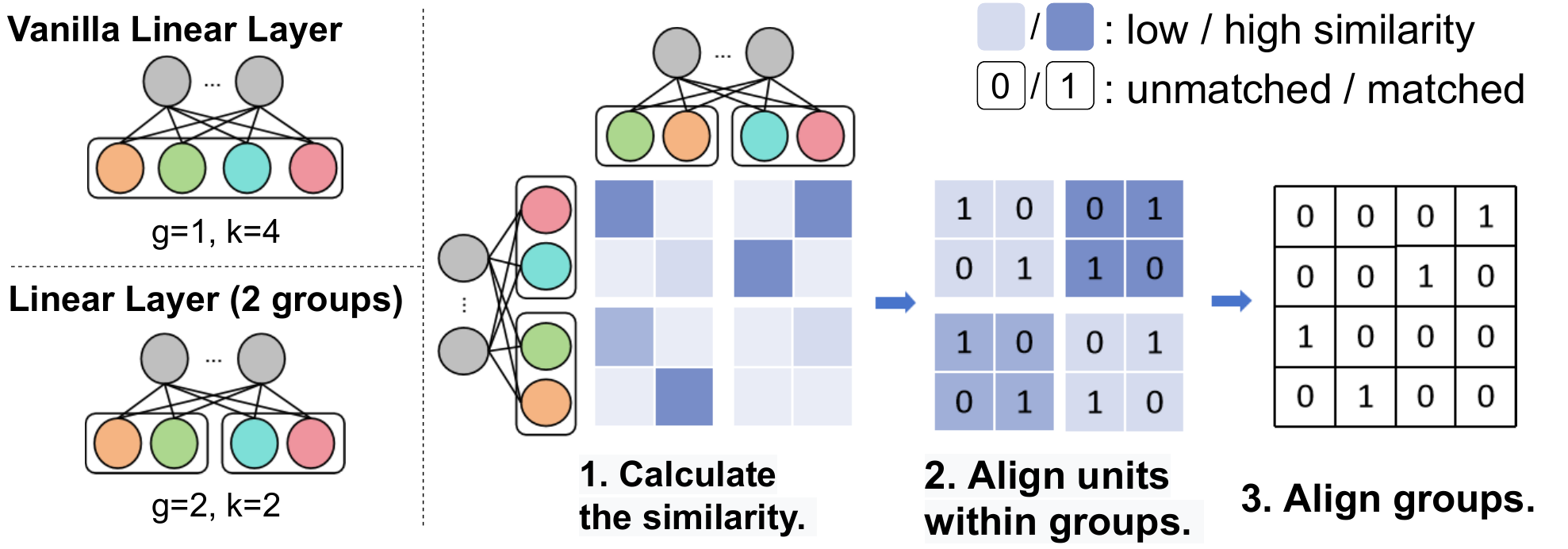}
   \caption{\textbf{Left.} Examples of vanilla structure and group structure.\textbf{Right.} An example of group alignment. First, we calculate the similarity between units. Next, we compute permutation and then
calculate the average of matched
similarity within each pairs of groups. Finally, we compute permutation for each pairs of groups and then set the permutation of unmatched pairs to zeros.}
   \label{fig:group_structure}
\end{figure}

Among them, group alignment first performs internal alignment of pairwise groups and then conducts external alignment based on the average similarity obtained from the internal alignment. We provide an illustrative diagram in Figure \ref{fig:group_structure}. While this algorithm has quadratic time complexity when it comes to external matching, we have found this cost to be acceptable for the application in our experiments. In addition, it is worth noting that this algorithm is capable of achieving a global optimal match.

For group zip, unlike alignment-based approaches, zip operation employs a greedy matching algorithm to merge similar units. However, this method is challenging to extend to group merging. In response to this, we propose a simplified matching scheme. We initially determine which groups to merge together through a group matching strategy, and then we "zip" these groups individually. As finding a suitable grouping strategy is a challenging task, we did not introduce a efficient grouping strategy in this work, leaving the task for future work. In subsequent experiments, we directly use the group matching results of the group alignment as the outcome of our grouping strategy, enabling a fair performance comparison between the alignment methods and the zip methods.

  



\section{Experiments}

We first validate the superiority of MuDSC to existing methods in merging models of homogenous tasks (\ie., classification tasks with different sets of categories) on both the small- and the large-scale datasets (Section~\ref{sec:split_class}). Then, we verify the effectiveness of MuDSC in merging models of heterogenous tasks (\eg., segmentation and classification) with the pre-trained models released in Taskonomy\cite{taskonomy_Zamir_2018_CVPR}~(Section~\ref{sec:rich_vision}). Finally, we visualize multi-task loss landscape to further demonstrate the superiority of MuDSC over prior single-space methods (Section \ref{sec:loss_landscape}). 


\begin{table*}
\centering
\begin{tabular}{c|cccc|cccc|cccc}
\toprule
\textbf{Model}      & \multicolumn{8}{c|}{\textbf{Resnet20}}                                                                                                 & \multicolumn{4}{c}{\textbf{Resnet20GN}}                           \\ \hline
\textbf{Dataset}    & \multicolumn{4}{c|}{\textbf{CIFAR100(50+50)}}                      & \multicolumn{4}{c|}{\textbf{CIFAR10(5+5)}}                         & \multicolumn{4}{c}{\textbf{CIFAR100(50+50)}}                      \\ \hline
\textbf{Method}     & \textbf{Joint} & \textbf{Avg}   & \textbf{T. A}  & \textbf{T. B}  & \textbf{Joint} & \textbf{Avg}   & \textbf{T. A}  & \textbf{T. B}  & \textbf{Joint} & \textbf{Avg}   & \textbf{T. A}  & \textbf{T. B}  \\ \midrule
\textbf{Average}    & 16.52          & 24.22          & 23.22          & 25.21          & 54.42          & 75.24          & 79.58          & 70.90          & 5.63           & 11.06          & 9.67           & 12.44          \\ \hline
\textbf{Rebasin}    & 41.33          & 56.94          & 57.31          & 56.58          & 60.61          & 88.57          & 88.46          & 88.68          & 13.85          & 22.18          & 22.99          & 21.37          \\
\textbf{A. Align}   & 44.33          & 61.13          & 61.61          & 60.66          & \textbf{61.71}          & 89.21          & 88.63          & \textbf{89.78}          & 29.37          & 42.05          & 41.05          & 43.05          \\
\textbf{MuDSC$_{Align}$} & \textbf{45.50} & \textbf{62.81} & \textbf{63.06} & \textbf{62.56} & 60.84          & \textbf{89.34} & \textbf{89.04} & 89.63          & \textbf{31.84} & \textbf{45.31} & \textbf{45.34} & \textbf{45.29} \\ \hline
\textbf{Zipit}      & 54.69          & 66.78          & 67.11          & 66.44          & 82.44          & 94.61          & 94.22          & 95.00          & 29.93          & 41.20          & 39.99          & 42.41          \\
\textbf{W.Zip}      & 55.16          & 67.65          & 68.58          & 66.71          & 82.85          & 94.71          & 94.42          & 94.99          & 14.28          & 20.95          & 19.17          & 22.72          \\
\textbf{MuDSC$_{Zip}$}   & \textbf{56.01} & \textbf{68.13} & \textbf{68.80} & \textbf{67.47} & \textbf{83.09} & \textbf{94.88} & \textbf{94.56} & \textbf{95.21} & \textbf{30.05}          & \textbf{41.52}          & \textbf{40.39} & \textbf{42.65}   \\
\bottomrule
\end{tabular}
  \caption{The joint accuracy and the per-task accuracy of the merged multitask model. Two original models are trained from scratch on two subtasks of the same classification task. We emphasize the data achieving the best accuracy. Std is provided in the supplementary material.}
  \label{tab:rd_init}
\end{table*}

\subsection{Merging Models of Homogeneous Tasks}

\label{sec:split_class}

\subsubsection{Experimental Settings}

\begin{table*}
\centering
\begin{tabular}{c|cccc|cccc|cccc}
\toprule
\textbf{Model}      & \multicolumn{4}{c|}{\textbf{Resnet26}}                             & \multicolumn{4}{c|}{\textbf{Resnet50GN}}                           & \multicolumn{4}{c}{\textbf{ViT}}                                  \\ \hline
\textbf{Method}     & \textbf{Joint} & \textbf{Avg}   & \textbf{T. A}  & \textbf{T. B}  & \textbf{Joint} & \textbf{Avg}   & \textbf{T. A}  & \textbf{T. B}  & \textbf{Joint} & \textbf{Avg}   & \textbf{T. A}  & \textbf{T. B}  \\
\midrule
\textbf{Average}    & 61.44          & 74.75          & 74.46          & 75.05          & 74.52          & 84.78          & 85.06          & 84.50          & 70.16          & 84.32          & 84.32          & 84.32          \\ \hline
\textbf{Rebasin}    & 61.39          & 74.79          & 74.48          & 75.10          & 74.52          & 84.78          & 85.06          & 84.50          & \textbf{70.16} & 84.32          & 84.32          & 84.32          \\
\textbf{A. Align}   & 61.91          & 75.41          & 75.03          & 75.79          & 74.44          & 84.77          & 84.99          & 84.56          & 69.99          & 84.22          & 84.20          & 84.24          \\
\textbf{MuDSC$_{Align}$} & \textbf{62.84} & \textbf{76.14} & \textbf{75.87} & \textbf{76.40} & \textbf{74.66} & \textbf{84.91} & \textbf{85.25} & \textbf{84.58} & 70.09          & \textbf{84.39} & \textbf{84.38} & \textbf{84.40} \\ \hline
\textbf{Zipit}      & 60.23          & 73.68          & 73.20          & 74.17          & 72.05          & 82.99          & 83.06          & 82.92          & 68.57          & 83.05          & 82.79          & 83.30          \\
\textbf{W.Zip}      & 61.28          & 74.69          & 74.42          & 74.96          & 74.52          & 84.78          & 85.06          & 84.50          & \textbf{70.16} & 84.32          & 84.32          & 84.32          \\
\textbf{MuDSC$_{Zip}$}   & \textbf{61.58} & \textbf{75.01} & \textbf{74.61} & \textbf{75.41} & \textbf{74.71} & \textbf{84.88} & \textbf{85.14} & \textbf{84.62} & 70.10          & \textbf{84.38} & \textbf{84.41} & \textbf{84.36} \\
\bottomrule
\end{tabular}
  \caption{The joint accuracy and the per-task accuracy of the merged multitask model. Two original models that well-pretrained on ImageNet are fintuned on two subtasks of CIFAR100. We emphasize the data achieving the best accuracy. Std is in the supplementary material.}
  \label{tab:finetuned}
\end{table*}

\begin{table}[t]
    \centering
\begin{tabular}{c|cccc}
\toprule
\textbf{Method}      & \textbf{Joint  Acc} & \textbf{Avg Acc}   & \textbf{T. A}  & \textbf{T. B}  \\ \midrule
\textbf{Average}     & 44.85          & 62.31          & 61.93          & 62.68          \\ \midrule
\textbf{Rebasin}     & 44.85          & 62.31          & 61.93          & 62.68          \\
\textbf{A. Align}    & 44.86          & 62.62          & 62.56          & 62.67          \\
\textbf{MuDSC$_{Align}$} & \textbf{44.87} & \textbf{62.66} & \textbf{62.57} & \textbf{62.74} \\ \midrule
\textbf{Zipit}       & 44.22          & 62.25          & 62.23          & 62.26          \\
\textbf{W.Zip}       & 44.85          & 62.31          & 61.93          & 62.68          \\
\textbf{MuDSC$_{Zip}$}   & \textbf{44.86} & \textbf{62.59} & \textbf{62.45} & \textbf{62.72}\\
\bottomrule
\end{tabular}
    \caption{Results on ImageNet400(200+200). Merging ResNet-50 models trained with DINO pretrained backbone on disjoint 200 category subsets (Task A and B) of ImageNet-1k.}
    \label{tab:imnet}
\end{table}

\noindent
\textbf{Datasets and models} The experiments are conducted on the small-scale CIFAR-10 \cite{cifar_krizhevsky2009learning}, CIFAR-100 \cite{cifar_krizhevsky2009learning} and the large-scale ImageNet~\cite{{imagenet}}. In the selection of models for merging, we emphasize the initial state of the model and whether it has a group structure \textit{i.e.} Multi-Head Attention \cite{vit_dosovitskiy2021an} and Group Normalization \cite{group_norm_Wu_2018_ECCV}. These two aspects of configuration significantly reflect the effectiveness and applicability of the merging algorithm. For randomly initialized models, ResNet-20 \cite{resnet_he2016deep} and ResNet-20 with Group Normalization \cite{group_norm_Wu_2018_ECCV} are selected. Additionally, we choose several pre-trained models from PyTorch Image Models (timm) \cite{timm_rw2019timm}, including ResNet-26 \cite{resnet_he2016deep}, ResNet-50 with Group Normalization\cite{group_norm_Wu_2018_ECCV}, ViT-Small \cite{vit_dosovitskiy2021an}, DINO-Small \cite{caron2021emerging} and Swin-Tiny \cite{liu2021swin}.


\noindent
\textbf{Baselines.} Our method is applied to two different types of matching algorithms referred to as alignment-based matching and zip-based matching. We compare our method with the current state-of-the-art methods of these two types. For alignment-based matching, the algorithm aligns the units of all models with those of the first model among them \cite{ainsworth2023git}. We implement a alignment-based version of our method MuDSC$_{Align}$ and compare it with Git Rebasin \cite{ainsworth2023git}. Additionally, we incorporated an activation-based alignment algorithm(abbreviated as A. Align). For zip-based matching, the units from different models are “zipped” together in the same layer which is referred to as zip operation \cite{stoica2023zipit}. Similarly, we implement MuDSC$_{Zip}$ based on zip operation, comparing it to Zipit \cite{stoica2023zipit}and a weight-based variant(abbreviated as W.Zip) implemented by us. 

\noindent
\textbf{Experimental details.} 
We randomly partition a classification dataset into two non-overlapping sub-classification tasks, trained respective models for each, and subsequently merged the models into one. Then we evaluate performance of merged model with joint accuracy and per-task accuracy. Joint accuracy is the overall accuracy of a model when it is evaluated on all classes within a combined dataset. It is similar to a continual learning setting where our objective is to enhance the knowledge of the model. For per-task accuracy, we provided the accuracy of the merged multi-task model on two individual tasks, along with their average performance. Each model is trained with a CLIP-style loss \cite{clip_radford2021learning} using CLIP text encodings of the class names as targets. For fair comparisons, we train 5 pairs of models and report the average accuracy. Three zip-based methods share the best hyperparameters (mentioned in \cite{stoica2023zipit}) on Zipit. In addition, for each pair of models, we search the balance factor $\alpha$ of MuDSC for the best per-task accuracy on the training data and then apply it to test.


\vspace{-0.5em}
\subsubsection{Results and Analysis}

\begin{table*}[t]
    \centering
\begin{tabular}{c|ccccccc}
\hline
                         & \multicolumn{7}{c}{\textbf{Method}}                                                                                                             \\ \cline{2-8} 
\textbf{Visual Task}              & \multicolumn{1}{c|}{\textbf{Weight Average}} & \textbf{Rebasin} & \textbf{Act. Align}       & \multicolumn{1}{c|}{\textbf{MuDSC$_{Align}$}}     & \textbf{Zipit}          & \textbf{Weight Zip} & \textbf{MuDSC$_{Zip}$}       \\ \hline
\textbf{Class Object}             & \multicolumn{1}{c|}{76.15}   & 80.76   & \textbf{89.75} & \multicolumn{1}{c|}{\textbf{89.75} $_{\textcolor{imp}{+0.00}}$} & 84.84          & 80.93  & \textbf{86.04} $_{\textcolor{imp}{+1.20}}$          \\
\textbf{Segment Semantic}         & \multicolumn{1}{c|}{23.01}   & 30.59   & 52.54          & \multicolumn{1}{c|}{\textbf{55.24} $_{\textcolor{imp}{+2.69}}$} & 32.51          & 33.77  & \textbf{36.22} $_{\textcolor{imp}{+2.45}}$          \\
\textbf{Rgb2depth}            & \multicolumn{1}{c|}{95.42}   & 95.90   & 98.69          & \multicolumn{1}{c|}{\textbf{98.70} $_{\textcolor{imp}{+0.01}}$} & 99.30          & 99.07  & \textbf{99.35} $_{\textcolor{imp}{+0.05}}$ \\
\textbf{Rgb2mist}          & \multicolumn{1}{c|}{94.56}   & 95.00   & \textbf{98.28} & \multicolumn{1}{c|}{\textbf{98.28} $_{\textcolor{imp}{+0.00}}$}          & 99.08          & 98.59  & \textbf{99.15} $_{\textcolor{imp}{+0.00}}$ \\
\textbf{Edge3D}                   & \multicolumn{1}{c|}{79.79}   & 80.44   & \textbf{91.04} & \multicolumn{1}{c|}{90.77 $_{\textcolor{dec}{-0.27}}$}          & \textbf{93.22} & 86.42  & 92.81 $_{\textcolor{dec}{-0.40}}$          \\
\textbf{Edge2D}                   & \multicolumn{1}{c|}{68.54}   & 75.48   & \textbf{81.37} & \multicolumn{1}{c|}{81.24 $_{\textcolor{dec}{-0.12}}$}          & 90.39          & 88.85  & \textbf{93.09} $_{\textcolor{imp}{+2.69}}$ \\
\textbf{Keypoints2D}              & \multicolumn{1}{c|}{71.33}   & 75.89   & 81.99          & \multicolumn{1}{c|}{\textbf{82.30} $_{\textcolor{imp}{+0.31}}$} & 93.38          & 93.62  & \textbf{94.73} $_{\textcolor{imp}{+1.11}}$ \\
\textbf{Keypoints3D}              & \multicolumn{1}{c|}{91.54}   & 91.99   & 96.16          & \multicolumn{1}{c|}{\textbf{96.17} $_{\textcolor{imp}{+0.02}}$} & \textbf{96.99} & 96.00  & 96.95 $_{\textcolor{dec}{-0.04}}$          \\
\textbf{Reshading}                & \multicolumn{1}{c|}{-4.72}   & 8.52    & 60.23          & \multicolumn{1}{c|}{\textbf{61.85} $_{\textcolor{imp}{+1.61}}$} & \textbf{69.18} & 41.66  & 68.71 $_{\textcolor{dec}{-0.47}}$          \\
\textbf{Rgb2sfnorm}           & \multicolumn{1}{c|}{26.62}   & 29.43   & 65.46          & \multicolumn{1}{c|}{\textbf{65.56} $_{\textcolor{imp}{+0.09}}$} & 76.62          & 64.43  & \textbf{77.50} $_{\textcolor{imp}{+0.88}}$ \\
\textbf{Autoencoding}             & \multicolumn{1}{c|}{21.96}   & 35.65   & 51.84          & \multicolumn{1}{c|}{\textbf{54.21} $_{\textcolor{imp}{+2.37}}$} & 86.55          & 77.58  & \textbf{87.99} $_{\textcolor{imp}{+1.45}}$ \\
\textbf{Denoising}                & \multicolumn{1}{c|}{33.92}   & 33.57   & 52.37          & \multicolumn{1}{c|}{\textbf{54.40} $_{\textcolor{imp}{+2.03}}$} & 84.28          & 76.14  & \textbf{85.49} $_{\textcolor{imp}{+1.22}}$ \\ \hline
\textbf{Total Average} & \multicolumn{1}{c|}{56.51}   & 61.10   & 76.64          & \multicolumn{1}{c|}{\textbf{77.37} $_{\textcolor{imp}{+0.73}}$} & 83.86          & 78.09  & \textbf{84.84} $_{\textcolor{imp}{+0.98}}$ \\ \hline
\end{tabular}
  \caption{The average scaled performance of multiple heterogenous-task models after merged with each other. We emphasize the data achieving the best accuracy and the improvement of our methods.}
  \label{tab:taskonomy}
\end{table*}

\noindent
\textbf{Results on CIFAR.} We first test the performance of the methods on merging randomly initialized models in Table \ref{tab:rd_init}. For experiments with Resnet20 on CIFAR100 and CIFAR10, the zip-based methods achieve relatively higher merged accuracy than alignment-based methods and our MuDSC$_{Zip}$ achieve further improvements. In experiments with Resnet20GN, MuDSC$_{Align}$ outperform all methods and significantly surpass the second-best method by 8.41\%, 7.75\%, 10.45\%, 5.20\% on joint accuracy and three per-task accuracy. Secondly, we conduct the experiments on CIFAR100 with Resnet26, Resnet50GN and ViT all of which are well-pretrained on ImageNet \cite{imagenet}. Unlike randomly initialized models, finetuned models exhibit a high similarity in weight space as they are initialized with the same pretrained model, making it challenging to discover better matching results. As observed in Table \ref{tab:finetuned}, the results for W. Zip and Rebasin are completely consistent with those of Average. Another regrettable fact is that when merging finetuned models, activation-based methods perform even worse than direct averaging. Fortunately, from Table \ref{tab:finetuned}, we can see that merging models under the dual-space constraints has overcome the obstacles of the aforementioned single-space methods, achieving further improvements in merged accuracy. Results of DINO-S and Swin-T (Table \ref{supplement-tab:more_vits}) also validate the superiority of our method.



\noindent
\textbf{Results on ImageNet.} To test our method on the large scale data, we train 5 ResNet-50 models initialized with DINO pretrained backbone \cite{caron2021emerging} on disjoint 200 class subsets of
ImageNet-1k \cite{imagenet}. Then we conduct experiments on exhaustively merging pairs from the 5 models. As shown in Tab. \ref{tab:imnet}, the proposed MuDSC still surpasses existing methods, demonstrating its capability for model merging on the large-scale data.

\subsection{Merging Models of Heterogenous Tasks}

\label{sec:rich_vision}

\subsubsection{Experimental Settings}

\noindent
\textbf{Tasks and models.} We adopt 12 pre-trained models from Taskonomy \cite{taskonomy_Zamir_2018_CVPR} trained on various tasks (including Autoencoder, Denoise, Edge 2D, Edge 3D, Keypoint 2D, Keypoint 3D, Reshade,
Rgb2depth, Rgb2mist, Rgb2sfnorm, Segmentation, and
Classification.). The architectures of these models follows an encoder-decoder scheme, in which the encoder is implemented by fully convolutional layers and the decoder varies according to the tasks. Please refer to \cite{taskonomy_Zamir_2018_CVPR} for more detailed information. In our experiments, only the encoders of these models are adopted for merging.


\noindent
\textbf{Baselines and Metric. }  The settings of baselines are the same as that of Sec. \ref{sec:split_class}. In order to reflect the gap between the merged model and the original model more intuitively, we define a scaled performance of model $\theta$, which is expressed as follows:

\begin{equation}
    \mathcal{L}_{SP}=\frac{\mathcal{L}_{\theta}-\mathcal{L}_{0}}{\mathcal{L}_{1}-\mathcal{L}_{0}},
\end{equation}
where $\mathcal{L}_{\theta}$, $\mathcal{L}_{1}$ and $\mathcal{L}_{0}$ represent the original performance metrics for the model $\theta$, the pretrained model and the average estimator. Notably, when $\mathcal{L}_{SP}$ approaches 1, the performance of the model $\theta$ is similar to the pre-trained model, and when $\mathcal{L}_{SP}$ approaches 0, the performance of the model $\theta$ is similar to the average value of the labels. In other words, we aim for the performance of the merged model to be closer to, or even better than 1. For the original performance, we maintain consistency with \cite{taskonomy_Zamir_2018_CVPR} except for adopting error rate in Segmentation, and we are unable to access any GAN loss due to the lack of discriminators.

\noindent
\textbf{Experimental Details.} We merge the adopted pretrained models with each other and ultimately report the average scaled performance per and all tasks for all merging methods. Due to the remarkably large size of the complete Taskonomy dataset, we conduct a reasonable sampling. For evaluation, we used the test split of official partitioned tiny subset, which contains 54513 samples, to evaluate the performance of the models. In order to reset the batch normalization of models and obtain the activations of the model, we sampled data from training dataset, which included 8 buildings and a total of 61520 samples. Additionally, the entire encoder is used to match but only the layers before the last layer are merged. In this experiment, we simply adopted 0.5 as the balanced factor for our MuDSC.


\begin{figure*}[t]
  \centering
  \includegraphics[width=1\textwidth]{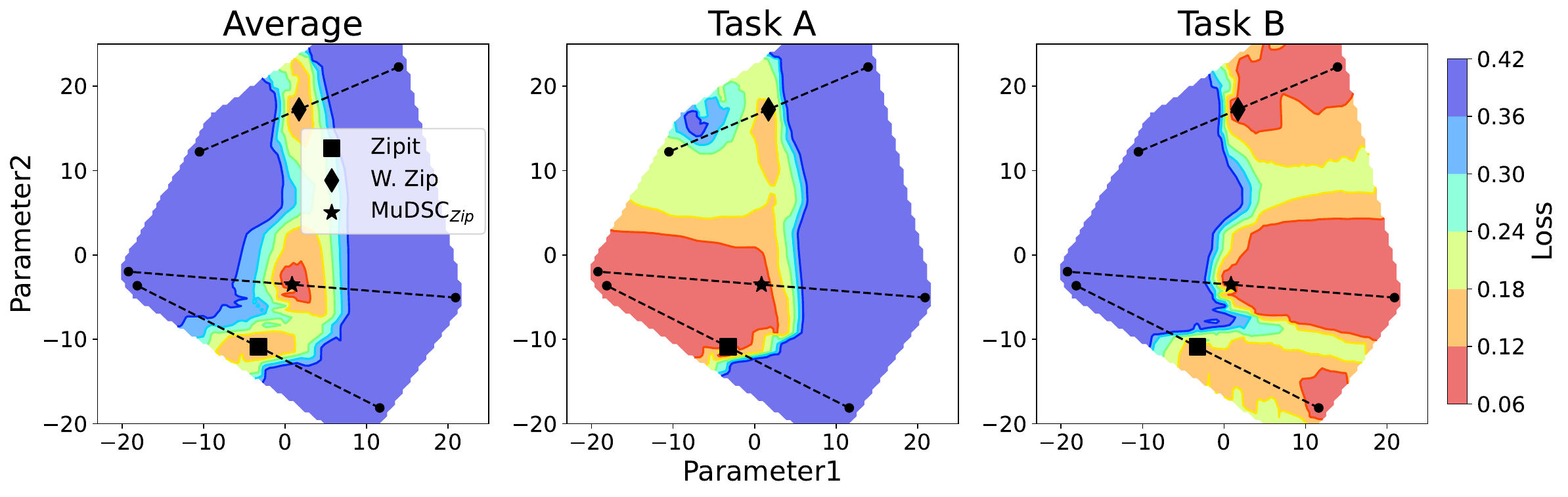}
   \caption{The loss landscape visualization of three zip-based methods(MuDSC$_{Zip}$, Zipit and W. Zip). Stars, squares, and diamond mark the positions of the merged models, and black spots mark the positions of their parent models.}
   \label{fig:loss_land}
\end{figure*}

\subsubsection{Results and Analysis}
Tab. \ref{tab:taskonomy} provide a quantitative comparison between the proposed MuDSC with existing works in scaled performance. It can be easily seen that the MuDSC yields significantly superior performance to existing two types of matching methods. In alignment-based algorithms, MuDSC$_{Align}$ maintains or surpasses the current SOTA in 83\% of tasks, 
Furthermore, in certain tasks \textit{i.e.}  Segment Semantic, Reshading, Autoencoding, and Denoise, MuDSC$_{Align}$ brings improvements of 5.12\%, 2.68\%, 4.58\%, and 3.88\%, respectively. Similarly, in zip-based methods, MuDSC outperforms the current SOTA methods in 75\% of tasks, with improvements exceeding 1\% in 8 tasks, notably achieving a remarkable 11.42\% improvement in the Segment Semantic task. Overall, alignment-based methods and zip-based methods have their respective strengths in merging tasks. Alignment-based methods excel at maintaining performance in semantic tasks(\textit{i.e.} Class Object and Segment Semantic), while zip-based methods have an advantage in pixel-to-pixel tasks. The reasons for this disparity are worth investigating further (but is out of role of our method). However, regardless of the merging methods, MuDSC brings significant improvements to both.

\subsection{Loss Landscape Visualization}

\label{sec:loss_landscape}

In this subsection, we demonstrate an example of how MuDSC improves the zip-based method. We conduct visualizations on a multi-task experiment constructed with MNIST \cite{lecun2010mnist}. Two binary classification tasks involve determining whether a sample is a prime number (\textit{i.e.} Task A) and whether a sample is odd (\textit{i.e.} Task B). The architecture of the model is a four-layer MLP and each model is trained as CLIP image encoder. The results are reported at Tab. \ref{supplement-tab:mt_mnist}. Then, we visualize merged models and their original models obtained by various methods onto a single diagram to intuitively compare their differences in the flatness of the loss landscape. 
To be more specific, we construct a set of high-dimensional vectors obtained by flattening models’ parameters and use the PCA dimension reduction algorithm \cite{pca_mackiewicz1993principal} to generate the two-dimensional coordinates. In Fig. \ref{fig:loss_land}, there are the average loss landscape, the loss landscape for Task A, and the loss landscape for Task B from left to right. Then we mark the positions of activation-based methods (\textit{i.e.} Zipit), weight-based methods, and MuDSC$_{Zip}$. 

As shown in Fig. \ref{fig:loss_land}, in the average loss landscape, our approach positions the merged model directly at the lowest point of the basin, in contrast to the activation-based method which places the merged model near the lowest point, and the weight-based method, which positions the merged model further away from the center of the basin. Then observe the loss landscape of different tasks. Upon further examination of the loss landscapes for different tasks, we observe that MuDSC facilitates the matching algorithm to discover a set of overlapping loss basins at lower points. This implies that our approach can better balance the performance of different tasks when merging for multi-task scenarios. Overall, with the support of a more precise constraint, MuDSC achieves an enhancement of the matching algorithm.

\section{Conclusion}
We introduce Merging under Dual-Space Constraints (MuDSC) to balance the inconsistency of unit similarity in weight space and activation space when merging models. MuDSC linearly combines the similarity matrices both of the weights and the activations of the units to seek a better permutation matrix. We find experimentally that MuDSC enhances the performance of the merged multi-task model across various tasks and architecture. We conduct the visualization of the merged model in the multi-task loss landscape which shows that MuDSC makes the merged model locate in the overlapping part with unified lower loss of each task. 

\section{Acknowledgements}

This work is supported by the Science and Technology Project of SGCC: Research and Digital Application of High-precision Electric Power Super-scale Pre-trained Visual Model (5108-202218280A-2-395-XG).


{
    \small
    \bibliographystyle{ieeenat_fullname}
    \bibliography{main}
}

\typeout{get arXiv to do 4 passes: Label(s) may have changed. Rerun}
\end{document}


\maketitle
\thispagestyle{empty}

\section{Details for Figure 1}
\label{sec:figure1}
For Figure 1, the experiments are conducted on ResNet-50. We first train it on CIFAR10 to get the first parent, and then retrain only the $5$-th convolution layer (64 units) to obtain the second parent. Then we average the two sets of units in the  $5$-th convolution layer pairwise, getting $64\times64=4,096$ merged units. The ``best merged" unit is the one whose similarity to the parents is maximized among the $4,096$ units. As each merged unit has two parents, "low bound" here means the smaller similarity value between the merged unit and its two parents. We use Pearson correlation to measure similarity.


\begin{table}[t]
\centering
\tabcolsep=0.48cm
\begin{tabular}{c|ccc}
    \toprule
\textbf{}     & \textbf{Avg Acc}          & \textbf{Prime} & \textbf{Odd}   \\ \midrule
\textbf{M. A}       & 71.71      & 98.81          & 44.60          \\
\textbf{M. B}        & 68.41      & 38.13          & 98.68          \\
\textbf{Avg}       & 63.99          & 58.61          & 69.37          \\ \hline
\textbf{Rebasin}     & 71.71         & 66.64          & 76.77          \\
\textbf{A. Align}       & 90.79        & 91.71          & 89.87          \\
\textbf{Ours$_{A.}$} & \textbf{90.93}  & \textbf{91.72} & \textbf{90.14} \\ \hline
\textbf{Zipit}       & 92.95        & 93.65          & 92.24          \\
\textbf{W. Zip}       & 89.35       & 87.90          & 90.79          \\
\textbf{Ours$_{Z.}$} & \textbf{94.62}   & \textbf{94.51} & \textbf{94.72} \\ \bottomrule
\end{tabular}
    \caption{The per-task accuracy on multi-task MNIST Dataset.}
    \label{tab:mt_mnist}
\end{table}

\section{Additional Results}

\noindent
\textbf{Original models and ensemble methods.} In Table \ref{tab:rd_init}, Table \ref{tab:finetuned} and Table \ref{tab:imnet}, we provide the results of original models and ensemble methods for the experiments in Section 4.1 as the reference.

\noindent
\textbf{Models used in Section 4.3.} The architecture of the model is a four-layer MLP, and each hidden layer has 1024 units. Each model is trained as CLIP image encoder. The per-task accuracy is shown in Table \ref{tab:mt_mnist}. We provide the average results of 5 different random seeds.

\begin{table}[h]
\centering
\begin{tabular}{c|cccc}
\toprule
\textbf{Method}      & \textbf{Joint  Acc} & \textbf{Avg Acc}   & \textbf{T. A}  & \textbf{T. B}  \\ \midrule
\textbf{Model A}     & 41.86          & 45.22          & 77.15          & 13.28          \\
\textbf{Model B}     & 40.81          & 45.14          & 13.30          & 76.98          \\
\textbf{Ensemble}    & 51.75          & 77.06          & 77.15          & 76.98          \\
\bottomrule
\end{tabular}
    \caption{Results of two original models and ensemble method for Table 3}
    \label{tab:imnet}
\end{table}

\begin{table*}[h!]
\centering
\tabcolsep=0.27cm
\begin{tabular}{c|cccc|cccc|cccc}
\toprule
\textbf{Model}      & \multicolumn{8}{c|}{\textbf{Resnet20}}                                                                                                 & \multicolumn{4}{c}{\textbf{Resnet20GN}}                           \\ \hline
\textbf{Dataset}    & \multicolumn{4}{c|}{\textbf{CIFAR100(50+50)}}                      & \multicolumn{4}{c|}{\textbf{CIFAR10(5+5)}}                         & \multicolumn{4}{c}{\textbf{CIFAR100(50+50)}}                      \\ \hline
\textbf{Method}     & \textbf{Joint} & \textbf{Avg}   & \textbf{T. A}  & \textbf{T. B}  & \textbf{Joint} & \textbf{Avg}   & \textbf{T. A}  & \textbf{T. B}  & \textbf{Joint} & \textbf{Avg}   & \textbf{T. A}  & \textbf{T. B}  \\ \midrule
\textbf{Model A}    & 41.48          & 53.44          & 82.71          & 24.18          & 48.59          & 70.98          & 96.67          & 45.29          & 38.70          & 49.16          & 77.11          & 21.22          \\
\textbf{Model B}    & 41.30          & 53.20          & 23.99          & 82.41          & 48.58          & 72.31          & 47.51          & 97.12          & 38.64          & 49.00          & 20.87          & 77.13          \\
\textbf{Ensemble}   & 69.51          & 82.56          & 82.71          & 82.41          & 84.12          & 96.89          & 96.67          & 97.12          & 63.18          & 77.12          & 77.10          & 77.13          \\
\bottomrule
\end{tabular}
  \caption{Results of original models and ensemble methods for Table 1}
  \label{tab:rd_init}
\end{table*}

\begin{table*}[h!]
\centering
{
\tabcolsep=0.27cm
\begin{tabular}{c|cccc|cccc|cccc}
\toprule
\textbf{Model}      & \multicolumn{4}{c|}{\textbf{Resnet26}}                             & \multicolumn{4}{c|}{\textbf{Resnet50GN}}                           & \multicolumn{4}{c}{\textbf{ViT}}                                  \\ \hline
\textbf{Method}     & \textbf{Joint} & \textbf{Avg}   & \textbf{T. A}  & \textbf{T. B}  & \textbf{Joint} & \textbf{Avg}   & \textbf{T. A}  & \textbf{T. B}  & \textbf{Joint} & \textbf{Avg}   & \textbf{T. A}  & \textbf{T. B}  \\
\midrule
\textbf{Model A}    & 42.89          & 54.31          & 84.72          & 23.89          & 45.02          & 56.99          & 89.24          & 24.75          & 47.57          & 58.10          & 93.05          & 23.14          \\
\textbf{Model B}    & 43.05          & 54.41          & 23.46          & 85.36          & 45.09          & 57.32          & 25.31          & 89.33          & 47.17          & 58.28          & 23.71          & 92.86          \\
\textbf{Ensemble}   & 71.43          & 85.04          & 84.72          & 85.36          & 76.88          & 89.28          & 89.23          & 89.32          & 82.69          & 92.95          & 93.05          & 92.86          \\
\bottomrule
\end{tabular}
}
  \caption{Results of original models and ensemble methods for Table 2}
  \label{tab:finetuned}
\end{table*}

\begin{table*}[h!]
\centering
\scalebox{0.9}{
\begin{tabular}{c|ccc|ccc|ccc}
\toprule
\textbf{Model}      & \multicolumn{6}{c|}{\textbf{Resnet20}}                                                                                                 & \multicolumn{3}{c}{\textbf{Resnet20GN}}                           \\ \hline
\textbf{Dataset}    & \multicolumn{3}{c|}{\textbf{CIFAR100(50+50)}}                      & \multicolumn{3}{c|}{\textbf{CIFAR10(5+5)}}                         & \multicolumn{3}{c}{\textbf{CIFAR100(50+50)}}                      \\ \hline
\textbf{Method}     & \textbf{Joint}    & \textbf{T. A}  & \textbf{T. B}  & \textbf{Joint}   & \textbf{T. A}  & \textbf{T. B}  & \textbf{Joint}    & \textbf{T. A}  & \textbf{T. B}  \\ \midrule
\textbf{Rebasin}    & 41.33$_{\pm 1.52}$                   & 57.31$_{\pm 1.23}$          & 56.58$_{\pm 0.28}$          & 60.61$_{\pm 0.14}$                   & 88.46$_{\pm 0.18}$          & 88.68$_{\pm 0.69}$          & 13.85$_{\pm 0.14}$                 & 22.99$_{\pm 0.36}$          & 21.37$_{\pm 0.42}$          \\
\textbf{A. Align}   & 44.33$_{\pm 0.13}$                    & 61.61$_{\pm 0.17}$          & 60.66$_{\pm 1.46}$          & \textbf{61.71}$_{\pm 0.13}$               & 88.63$_{\pm 0.06}$          & \textbf{89.78}$_{\pm 0.52}$          & 29.37$_{\pm 1.07}$                & 41.05$_{\pm 1.44}$          & 43.05$_{\pm 0.85}$          \\
\textbf{MuDSC$_{Align}$} & \textbf{45.50}$_{\pm 0.38}$  & \textbf{63.06}$_{\pm 0.46}$ & \textbf{62.56}$_{\pm 0.48}$ & 60.84$_{\pm 0.14}$    & \textbf{89.04}$_{\pm 0.25}$ & 89.63$_{\pm 0.21}$          & \textbf{31.84}$_{\pm 0.60}$  & \textbf{45.34}$_{\pm 1.17}$ & \textbf{45.29}$_{\pm 0.79}$ \\ \hline
\textbf{Zipit}      & 54.69$_{\pm 0.15}$                    & 67.11$_{\pm 0.77}$          & 66.44$_{\pm 0.68}$          & 82.44$_{\pm 0.76}$                & 94.22$_{\pm 0.14}$          & 95.00$_{\pm 0.95}$          & 29.93$_{\pm 1.09}$                   & 39.99$_{\pm 0.73}$          & 42.41$_{\pm 0.86}$          \\
\textbf{W.Zip}      & 55.16$_{\pm 0.20}$                & 68.58$_{\pm 0.29}$          & 66.71$_{\pm 0.08}$          & 82.85$_{\pm 0.13}$                & 94.42$_{\pm 0.15}$          & 94.99$_{\pm 0.11}$          & 14.28$_{\pm 1.07}$                   & 19.17$_{\pm 0.78}$          & 22.72$_{\pm 0.88}$          \\
\textbf{MuDSC$_{Zip}$}   & \textbf{56.01}$_{\pm 0.25}$  & \textbf{68.80}$_{\pm 0.05}$ & \textbf{67.47}$_{\pm 0.33}$ & \textbf{83.09}$_{\pm 0.13}$  & \textbf{94.56}$_{\pm 0.24}$ & \textbf{95.21}$_{\pm 0.39}$ & \textbf{30.05}$_{\pm 0.39}$                  & \textbf{40.39}$_{\pm 0.70}$ & \textbf{42.65}$_{\pm 0.56}$   \\
\bottomrule
\end{tabular}
}
  \caption{Results of MuDSC in Table 1 including std.}
  \label{tab:rd_init_std}
\end{table*}

\begin{table*}[h!]
\centering
\scalebox{0.9}{
\begin{tabular}{c|ccc|ccc|ccc}
\toprule
\textbf{Model}      & \multicolumn{3}{c|}{\textbf{Resnet26}}                             & \multicolumn{3}{c|}{\textbf{Resnet50GN}}                           & \multicolumn{3}{c}{\textbf{ViT}}                                  \\ \hline
\textbf{Method}     & \textbf{Joint}   & \textbf{T. A}  & \textbf{T. B}  & \textbf{Joint}   & \textbf{T. A}  & \textbf{T. B}  & \textbf{Joint}   & \textbf{T. A}  & \textbf{T. B}  \\
\midrule
\textbf{Rebasin}    & 61.39$_{\pm 0.31}$               & 74.48$_{\pm 0.36}$          & 75.10$_{\pm 0.18}$          & 74.52$_{\pm 0.12}$                   & 85.06$_{\pm 0.32}$          & 84.50$_{\pm 0.13}$          & \textbf{70.16}$_{\pm 0.15}$           & 84.32$_{\pm 0.05}$          & 84.32$_{\pm 0.02}$          \\
\textbf{A. Align}   & 61.91$_{\pm 0.44}$               & 75.03$_{\pm 0.43}$          & 75.79$_{\pm 0.34}$          & 74.44$_{\pm 0.13}$                   & 84.99$_{\pm 0.04}$          & 84.56$_{\pm 0.01}$          & 69.99$_{\pm 0.16}$                   & 84.20$_{\pm 0.07}$          & 84.24$_{\pm 0.12}$          \\
\textbf{MuDSC$_{Align}$} & \textbf{62.84}$_{\pm 0.50}$  & \textbf{75.87}$_{\pm 0.38}$ & \textbf{76.40}$_{\pm 0.35}$ & \textbf{74.66}$_{\pm 0.09}$ & \textbf{85.25}$_{\pm 0.07}$ & \textbf{84.58}$_{\pm 0.03}$ & 70.09$_{\pm 0.03}$           & \textbf{84.38}$_{\pm 0.13}$ & \textbf{84.40}$_{\pm 0.04}$ \\ \hline
\textbf{Zipit}      & 60.23$_{\pm 0.70}$                  & 73.20$_{\pm 0.89}$          & 74.17$_{\pm 0.71}$          & 72.05$_{\pm 0.52}$                    & 83.06$_{\pm 0.29}$          & 82.92$_{\pm 0.12}$          & 68.57$_{\pm 0.16}$                   & 82.79$_{\pm 0.12}$          & 83.30$_{\pm 0.18}$          \\
\textbf{W.Zip}      & 61.28$_{\pm 0.06}$                    & 74.42$_{\pm 0.23}$     & 74.96$_{\pm 0.08}$             & 74.52$_{\pm 0.12}$                    & 85.06$_{\pm 0.32}$          & 84.50$_{\pm 0.13}$          & \textbf{70.16}$_{\pm 0.15}$           & 84.32$_{\pm 0.05}$    & 84.32$_{\pm 0.02}$             \\
\textbf{MuDSC$_{Zip}$}   & \textbf{61.58}$_{\pm 0.27}$  & \textbf{74.61}$_{\pm 0.24}$ & \textbf{75.41}$_{\pm 0.38}$ & \textbf{74.71}$_{\pm 0.01}$  & \textbf{85.14}$_{\pm 0.01}$ & \textbf{84.62}$_{\pm 0.03}$ & 70.10$_{\pm 0.03}$        & \textbf{84.41}$_{\pm 0.08}$ & \textbf{84.36}$_{\pm 0.05}$ \\
\bottomrule
\end{tabular}
}
  \caption{Results of MuDSC in Table 2 including std.}
  \label{tab:finetuned_std}
\end{table*}

\noindent
\textbf{More advanced ViTs.} Here we provide the results on DINO \cite{caron2021emerging} and Swin-Transformer \cite{liu2021swin} in Table \ref{tab:more_vits}. It can seen that the proposed method consistently outperforms the two competitors, again validate the superiority of the our method.

\noindent
\textbf{Statistical significance.} Here we provide some stds of our methods and the two competitors in Table \ref{tab:rd_init_std} and Table \ref{tab:finetuned_std}, which proves the significance of the improvements.

\begin{table*}[h!]
\centering
{
\tabcolsep=0.5cm
\begin{tabular}{c|ccc|ccc}
\toprule
\textbf{Model}    & \multicolumn{3}{c|}{\textbf{DINO-S}} & \multicolumn{3}{c}{\textbf{Swin-T}} \\ \hline
\textbf{Method}   & \textbf{Joint}   & \textbf{T. A}    & \textbf{T. B}   & \textbf{Joint}   & \textbf{T. A}    & \textbf{T. B}   \\ \midrule
\textbf{Rebasin}  & $66.22_{\pm 0.25}$   & $81.07_{\pm 0.10}$   & $78.37_{\pm 0.43}$  & $75.41_{\pm 0.08}$   & $87.46_{\pm 0.16}$   & $84.99_{\pm 0.29}$  \\
A. Align & $61.28_{\pm 1.73}$   & $76.92_{\pm 1.33}$   & $74.46_{\pm 1.61}$  & $67.86_{\pm 0.56}$   & $81.42_{\pm 0.57}$   & $79.12_{\pm 0.47}$  \\
\textbf{MuDSC A.} & $\textbf{66.25}_{\pm 0.25}$   & $\textbf{81.24}_{\pm 0.10}$   & $\textbf{79.20}_{\pm 0.5}$  & $\textbf{75.76}_{\pm 0.12}$   & $\textbf{87.78}_{\pm 0.16}$   & $\textbf{85.56}_{\pm 0.32}$  \\ \hline
\textbf{Zipit}    & $59.73_{\pm 0.22}$   & $79.90_{\pm 0.15}$   & $74.24_{\pm 0.56}$  & $63.39_{\pm 0.81}$   & $75.22_{\pm 1.22}$   & $73.09_{\pm 0.75}$  \\
\textbf{W. Zip}   & $66.21_{\pm 0.25}$   & $81.11_{\pm 0.10}$   & $78.37_{\pm 0.44}$  & $75.41_{\pm 0.08}$   & $87.46_{\pm 0.16}$   & $84.99_{\pm 0.29}$  \\
\textbf{MuDSC Z.} & $\textbf{66.66}_{\pm 0.29}$   & $\textbf{81.27}_{\pm 0.09}$   & $\textbf{79.35}_{\pm 0.48}$  & $\textbf{75.73}_{\pm 0.10}$   & $\textbf{87.74}_{\pm 0.15}$   & $\textbf{85.52}_{\pm 0.33}$ \\ \bottomrule
\end{tabular}
}
  \caption{CIFAR100 Results on DINO and Swin-Transformer.}
  \label{tab:more_vits}

\end{table*}

\section{Convergence of MuDSC}
\label{sec:convergence}
Algorithm 1 adopts a well-proved iterative algorithm \cite{ainsworth2023git}, where each iteration increases the similarity until it converges.

\section{Complexity of MuDSC}
\label{sec:complexity}
As solving Eq. 1 dominates the computation of Alg. 1, here we simply discuss the complexity of solving Eq. 1. The activation-based methods are single-round methods as they can solve Eq. 1 in just one round, while weight-based methods and the proposed MuDSC are multi-round methods as they solve Eq. 1 in an iterative manner. Tab. \ref{tab:gpu_time} provides some results. It can be seen that MuDSCs needs less rounds than prior weight-based methods if two models are trained from scratch.
\begin{table*}[t]
\centering
\scalebox{0.95}{
\begin{tabular}{c|c|ccc|c|ccc}
\toprule
\multirow{2}{*}{} & \multirow{2}{*}{\textbf{Align Time/Round}} & \multicolumn{3}{c|}{\#~\textbf{Rounds}} & \multirow{2}{*}{\textbf{Zip Time/Round}} & \multicolumn{3}{c}{\#~\textbf{Rounds}} \\ 
                  &                                   & \textbf{A. Align}    & \textbf{Rebasin}    & \textbf{MuDSC A.}   &                                 & \textbf{Zipit}     & \textbf{W. Align}    & \textbf{MuDSC Z.}    \\ \midrule
\textbf{Resnet20}          & 0.12sec                           & 1           & 10         & 5          & 2.76sec                         & 1         & 6           & 5           \\
\textbf{Resnet26(pre)}     & 0.42sec                           & 1           & 4          & 3          & 16.12sec                        & 1         & 3           & 3           \\
\textbf{ViT-S(pre)}        & 0.97sec                           & 1           & 2          & 3          & 1.02min                         & 1         & 3           & 3     \\
\bottomrule
\end{tabular}
}
  \caption{Time to solve Eq. 1 and the rounds to converge.}
  \label{tab:gpu_time}
\end{table*}

    


{
    \small
    \bibliographystyle{ieeenat_fullname}
    \bibliography{main}
}

\typeout{get arXiv to do 4 passes: Label(s) may have changed. Rerun}

%% file: preamble.tex
\usepackage[dvipsnames]{xcolor}
